\documentclass[runningheads]{llncs}

\usepackage[utf8]{inputenc} 
\usepackage[T1]{fontenc}    
\usepackage{hyperref}       
\usepackage{url}            
\usepackage{booktabs}       
\usepackage{amsfonts}       
\usepackage{nicefrac}       
\usepackage{microtype}      
\usepackage{xcolor}         
\usepackage{amsmath}
\usepackage{graphicx}
\usepackage{rotating}
\usepackage{tabulary}
\usepackage{algorithm}
\usepackage{algorithmic}
\usepackage{subdef}
\usepackage{wrapfig,lipsum,booktabs}
\usepackage{subcaption}
\usepackage[numbers]{natbib}

\newcommand{\figref}[1]{Fig.~\ref{#1}}
\newcommand{\tabref}[1]{Tab.~\ref{#1}}
\newcommand{\secref}[1]{Sec.~\ref{#1}}
\newcommand{\AlgRef}[1]{Algo.~\ref{#1}}

\newcommand{\Appref}[1]{Appendix.~\ref{#1}}
\newcommand{\model}{\mbox{\textsc{Basil}}}
\begin{document}

\title{BASIL: Balanced Active Semi-supervised Learning for Class Imbalanced Datasets}
\titlerunning{BASIL: Balanced Active Semi-supervised Learning}
%
\author{Suraj Kothawade\inst{1} \and
Pavan Reddy\inst{2} \and
Ganesh Ramakrishnan\inst{2}\and
Rishabh Iyer\inst{1}}
\authorrunning{S. Kothawade et al.}
%
\institute{University of Texas at Dallas, USA \and
Indian Institute of Technology, Bombay, India\\
\email{suraj.kothawade@utdallas.edu}}

\maketitle              

\begin{abstract}
Current semi-supervised learning (SSL) methods assume a balance between the number of data points available for each class in both the labeled and the unlabeled data sets. However, there naturally exists a class imbalance in most real-world datasets. It is known that training models on such imbalanced datasets leads to biased models, which in turn lead to biased predictions towards the more frequent classes. This issue is further pronounced in SSL methods, as they would use this biased model to obtain psuedo-labels (on the unlabeled data) during training. In this paper, we tackle this problem by attempting to select a balanced labeled dataset for SSL that would result in an unbiased model. Unfortunately, acquiring a balanced labeled dataset from a class imbalanced distribution in one shot is challenging. We propose \model\ (\textbf{B}alanced \textbf{A}ctive \textbf{S}emi-superv\textbf{I}sed \textbf{L}earning), a novel algorithm that optimizes the submodular mutual information (SMI) functions in a per-class fashion to gradually select a balanced dataset in an active learning loop. Importantly, our technique can be efficiently used to improve the performance of any SSL method. Our experiments on Path-MNIST and Organ-MNIST medical datasets for a wide array of SSL methods show the effectiveness of \model. Furthermore, we observe that \model\ outperforms the state-of-the-art diversity and uncertainty based active learning methods since the SMI functions select a more balanced dataset.  

\end{abstract}

\section{Introduction}
Deep neural networks (DNNs) have proven to be successful on a variety of machine learning tasks. However, they are mostly fueled by large amounts of data. These DNNs can be trained using multiple objective functions that require labeled or unlabeled data. Towards this end, we face multiple data related challenges in training DNNs. \textbf{Firstly,} most real-world datasets have a natural class-imbalance. For instance, in the medical imaging domain, cancerous images are rare in comparison to non-cancerous images. \textbf{Secondly,} obtaining labeled data is notoriously time-consuming and expensive. This issue is highly pronounced in the biomedical domain where the annotators need to be well compensated since they are experts like doctors, radiologists, \etc\ \textbf{Thirdly,} in many scenarios, even procuring unlabeled data is challenging. For example, acquiring few samples of medical data of a new disease is rare and involves several privacy constraints. Hence, it is critical to use the unlabeled data even when a few labeled data points are available. \looseness-1

The above introduced data related issues are well-known, and the community has devised several techniques to tackle these issues \emph{seperately}. For mitigating labeling costs, \emph{active learning} (AL)~\cite{ash2020deep, kothawade2021talisman, kirsch2019batchbald, kothawade2021similar, sener2018active, settles2009active} is an established paradigm that samples uncertain or diverse data points from an unlabeled set. The goal is to acquire a subset that entails the largest improvement in performance of the model. Another technique, called \emph{semi-supervised learning} (SSL)~\cite{lee2013pseudo, miyato2018virtual, laine2016temporal, berthelot2019mixmatch, tarvainen2017mean, verma2019interpolation} leverages the unlabeled data when only a small amount of labeled data is available. Lastly, several subset selection techniques that tackle class imbalance~\cite{kothawade2021similar, kothawade2021prism, kothawade2021talisman} have also been proposed. Evidently, these techniques revolve around the idea of obtaining the best possible model at a minimum cost. However, individually, each of these techniques suffer from limitations that the other ones do not. For example, existing AL and SSL techniques are known to suffer from class imbalance, thereby leading to learning biased models. Also, existing AL and subset selection methods do not leverage the remaining unlabeled data, and simply discard it. To bridge these gaps in existing methods, we propose \model, a unified framework that actively samples data points per-class to create a balanced labeled set followed by SSL to make the most of the remaining unlabeled data. \looseness-1

\subsection{Related work} \label{sec:related_work}

\textbf{Active Learning (AL).} Uncertainty based methods aim to select the most uncertain data points for labeling. The most common technique is \textsc{Entropy}~\cite{settles2009active} that aims to select data points with maximum entropy. The main drawback of uncertainty based methods is that they lack diversity within the acquired subset. To mitigate this, a number of approaches have proposed to incorporate diversity. A recent approach called \textsc{Badge}~\cite{ash2020deep} uses the last linear layer gradients to represent data points and runs \textsc{K-means++}~\cite{kmeansplus} to obtain centers that have a high gradient magnitude. The centers being representative and having high gradient magnitude ensures uncertainty and diversity at the same time. However, for batch active learning, this diversity and uncertainty are limited within the batch and \emph{not} across all batches. Another method, \textsc{BatchBald}~\cite{kirsch2019batchbald} requires a large number of Monte Carlo dropout samples to obtain significant mutual information which limits its application to medical domains where data is scarce. Recently, \cite{kothawade2021similar} proposed the use of submodular information measures for active learning in realistic scenarios, while~\cite{kothawade2021talisman} used them to find rare objects in an autonomous driving object detection problem. However, they focus on acquiring data points \emph{only} from the rare classes or slices. Our proposed method maximizes the per-class mutual information, thereby selecting data points for \emph{each} class to obtain a balanced labeled set, which is critical for training unbiased models.

\vspace{0.2cm}
\noindent \textbf{Semi-supervised Learning (SSL). } The goal of SSL methods is to leverage unlabeled data alongside the labeled data to obtain a better representation of the dataset than supervised learning~\cite{oliver2018realistic}. The most basic SSL method, pseudo-labeling~\cite{lee2013pseudo} uses model predictions as target labels as a regularizer, and a standard supervised loss function for the unlabeled dataset. Some SSL methods such as $\Pi$-Model~\cite{laine2016temporal, sajjadi2016regularization} and Mean Teacher~\cite{tarvainen2017mean} use consistency regularization, by using data augmentation and dropout techniques. Mean Teacher obtains a more stable target output by using an exponential moving average of parameters across previous epochs. Virtual Adversarial Training (VAT)~\cite{miyato2018virtual} uses an effective regularization technique that involves slight perturbations such that the prediction of the unlabeled samples is affected the most. ICT~\cite{verma2019interpolation} encourages the prediction at an interpolation of unlabeled points to be consistent with the interpolation of the predictions at those points. More recent techniques like FixMatch~\cite{sohn2020fixmatch}, MixMatch~\cite{berthelot2019mixmatch} and UDA~\cite{xie2019unsupervised} use data augmentations like flip, rotation, and crops to predict pseudo-labels. All the above methods depend on the model trained using a small labeled set. Hence, they are susceptible to using a biased model if the labeled set is randomly sampled from an unlabeled set with class imbalance. In this paper, we study the effect of selecting a balanced seed set using \model\ for a wide array of SSL techniques.

\subsection{Our contributions} \label{sec:contributions}
We summarize our contributions as follows: \textbf{1)} We emphasize on the need of jointly addressing multiple real-world data related problems such as class imbalance, expensive labeling costs, and leveraging the unlabeled data. Particularly, we show that these problems \emph{co-exist} in the medical domain. \textbf{2)} We propose \model, a novel algorithm that can tackle these problems in an end-to-end manner. Concretely, we acquire a balanced subset of the unlabeled data by maximizing per-class instantiations of submodular mutual information functions in an active learning loop followed by semi-supervised learning (see \figref{fig:basil_arch}). \textbf{3)} \model\ can leverage any SSL method and yield improved performance over the vanilla SSL approach. \textbf{4)} We evaluate the effectiveness of \model\ on two diverse modalities of medical data. Namely, histopathology (Path-MNIST~\cite{kather2019predicting}) and Abdominal CT (Organ-MNIST~\cite{kermany2018identifying}). \textbf{5)} We conduct rigorous experiments on 6 AL strategies and 8 SSL techniques, and show that balanced labeled set selection using \model\ outperforms existing AL methods and obtains larger gain in performance for various SSL techniques (see~\tabref{tab:pathmnist_res} and \tabref{tab:organmnist_res}).

\section{Preliminaries} \label{sec:preliminaries}

\textbf{Submodular Functions: } We let $\Vcal$ denote the \emph{ground-set} of $n$ data points $\Vcal = \{1, 2, 3,...,n \}$ and a set function $f:
 2^{\Vcal} \xrightarrow{} \mathbb{R}$.  
 The function $f$ is submodular~\citep{fujishige2005submodular}  if it satisfies the diminishing marginal returns, namely $f(j | \Xcal) \geq f(j | \Ycal)$ for all $\Xcal \subseteq \Ycal \subseteq \Vcal, j \notin \Ycal$.
Facility location, graph cut, log determinants, {\em etc.} are some examples~\citep{iyer2015submodular}. 

\noindent \textbf{Submodular Mutual Information (\textsc{Smi}):} Given a set of items $\Acal, \Qcal \subseteq \Vcal$, the submodular mutual information (MI)~\citep{levin2020online,iyer2020submodular} is defined as $I_f(\Acal; \Qcal) = f(\Acal) + f(\Qcal) - f(\Acal \cup \Qcal)$. Intuitively, this measures the similarity between $\Qcal$ and $\Acal$ and we refer to $\Qcal$ as the query set.

\noindent {\em Kothawade et. al.}~\cite{kothawade2021prism} extend \textsc{Smi} to handle the case when the \emph{target} can come from a different set $\Vcal'$ apart from the ground set $\Vcal$. In the context of imbalanced medical image classification, $\Vcal$ is the source set of images and the query set $\Qcal$ is the target set containing the rare class images.
To find an optimal subset given a query set $\Qcal \subseteq \Vcal^{\prime}$, we can define $g_{\Qcal}(\Acal) = I_f(\Acal; \Qcal)$, $\Acal \subseteq \Vcal$ and maximize the same. \looseness-1

\subsection{Examples of \textsc{Smi} functions}
For balanced subset selection via \model\, we use the recently introduced \textsc{Smi} functions in~\cite{iyer2020submodular, levin2020online} and their extensions introduced in \cite{kothawade2021prism} as acquisition functions. Note that we only use a subset of functions presented in \cite{kothawade2021prism}, that are the most scalable, for per-class selection of data points.
For any two data points $i \in \Vcal$ and $j \in \Qcal$, let $s_{ij}$ denote the similarity between them.

\noindent\textbf{Graph Cut MI (\textsc{Gcmi}):} The \textsc{Smi} instantiation of graph-cut (\textsc{Gcmi}) is defined as: $I_{GC}(\Acal;\Qcal)=2\sum_{i \in \Acal} \sum_{j \in \Qcal} s_{ij}$.
Since maximizing \textsc{Gcmi} maximizes the joint pairwise sum with the query set, it will lead to a summary similar to the query set $Q$. In fact, specific instantiations of \textsc{Gcmi} have been intuitively used for query-focused summarization for videos~\cite{vasudevan2017query} and documents~\cite{lin2012submodularity, li2012multi}. 


\noindent\textbf{Facility Location MI (\textsc{Flmi}):} We consider two variants of \textsc{Flmi}. In the first variant of facility location which is defined over $\Vcal$(\textsc{Flvmi}), the \textsc{Smi} instantiation can be defined as: $I_{FLV}(\Acal;\Qcal)=\sum_{i \in \Vcal}\min(\max_{j \in \Acal}s_{ij}, \max_{j \in \Qcal}s_{ij})$.
The first term in the min(.) of \textsc{Flvmi} models diversity, and the second term models query relevance. 

\noindent For the second variant, which is defined over $\Qcal$ (\textsc{Flqmi}), the \textsc{Smi} instantiation can be defined as: $I_{FLQ}(\Acal;\Qcal)=\sum_{i \in \Qcal} \max_{j \in \Acal} s_{ij} + \sum_{i \in \Acal} \max_{j \in \Qcal} s_{ij}$. \textsc{Flqmi} is very intuitive for query relevance as well. It measures the representation of data points that are the most relevant to the query set and vice versa. It can also be thought of as a bidirectional representation score.

\begin{figure*}
\includegraphics[width = \textwidth]{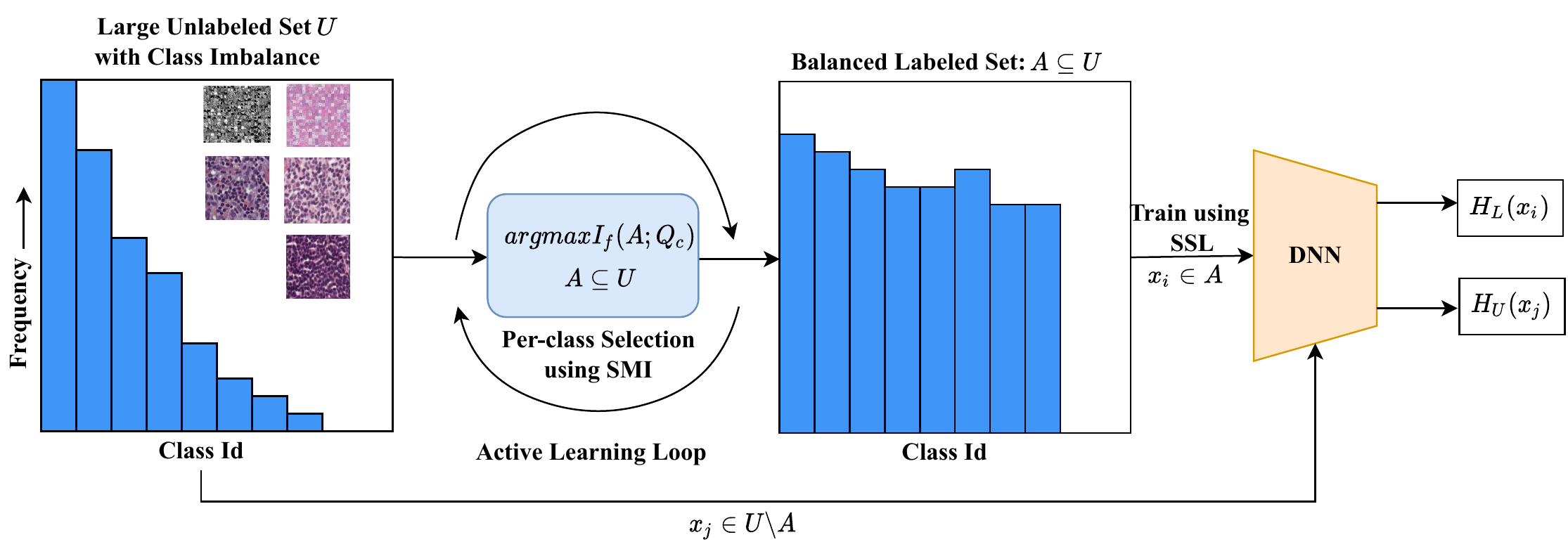}
\caption{Architecture of \model. We select a balanced labeled set $\Acal \subseteq \Ucal$ by maximizing the per-class submodular mutual information in an active learning loop. The balanced labeled set is used to train an unbiased model which better leverages the remaining unlabeled data $\Ucal \backslash \Acal$ using semi-supervised learning.}
\label{fig:basil_arch}
\end{figure*}
\vspace{-6ex}
 
\section{\model: Our Active Semi-supervised Learning framework} \label{sec:our_method}
In this section, we present \model, a unified framework that jointly tackles data problems of class imbalance, high labeling costs and leveraging unlabeled data. We do so by gradually acquiring balanced subsets in an active learning loop, followed by training the final model using the balanced labeled set and the remaining unlabeled set (see \figref{fig:basil_arch}).

\begin{algorithm}
\begin{algorithmic}[1]
\REQUIRE Labeled set of data points: $\Lcal$, large unlabeled dataset: $\Ucal$, Loss function defined over $\Lcal$: $\Hcal_\Lcal$, Loss function defined over $\Ucal$: $\Hcal_\Ucal$, batch size: $B$, number of selection rounds: $N$, number of classes: $C$ \\
\FOR{selection round $i = 1:N$}
\STATE Train model $\Mcal_{\theta_i}$ with loss $\Hcal_\Lcal$ on the current labeled set $\Lcal$
\STATE $\Gcal_\Ucal \leftarrow  \{\nabla_{\theta_i} \mathcal H_\Lcal(x_j, \hat{y_j}, \theta_i), \forall j \in \Ucal\}$ \textcolor{blue}{\{Compute gradients of $\Ucal$ using hypothesized labels\}}
\STATE $\Gcal_\Lcal \leftarrow \{\nabla_{\theta_i} \mathcal H_\Lcal(x_j, y_j, \theta_i), \forall j \in \Lcal\}$ \textcolor{blue}{\{Compute gradients of $\Lcal$ using true labels\}}
\STATE $\Xcal \leftarrow$ \textsc{Cosine\_Similarity} ($\Gcal_\Lcal, \Gcal_\Ucal$) \textcolor{blue}{\{$X \in \mathbb{R}^{|\Lcal| \times |\Ucal|}$\}}
\FOR{class $c \in \{1 \cdots C\}$} 
\STATE Instantiate a \textsc{Smi} function $I_f^c$ based on $\Xcal^c$. \textcolor{blue}{\{$\Xcal^c$ contains rows from $\Xcal$ corresponding to class $c$\}}
\STATE $\Acal_i^c \leftarrow \mbox{argmax}_{\Acal_i^c \subseteq \Ucal, |\Acal_i^c| \leq (B/C)}  I_f(\Acal_i^c; \Qcal^c)$ \textcolor{blue}{\{$\Qcal^c \subseteq \Lcal$ contains data points from class $c$\}}
\STATE $\Acal_i \leftarrow \Acal_i \cup \Acal_i^c$
\ENDFOR
\STATE Get labels $L(\Acal_i)$ for batch $\Acal_i$ and $\Lcal \leftarrow \Lcal \cup L(\Acal_i)$, $\Ucal \leftarrow \Ucal - \Acal_i$
\ENDFOR
\STATE Train final model $\Mcal_\theta$ with loss $\Hcal_\Lcal + \Hcal_{\Ucal}$ \textcolor{blue}{\{Train model on balanced $\Lcal$ and remaining $\Ucal$ using SSL\}}
\STATE Return trained model $\Mcal$ and parameters $\theta$.
\end{algorithmic}
\caption{\model: Balanced Active Semi-supervised Learning}
\label{algo:basil}
\end{algorithm}

The main idea in \model\ is to maximize per-class instantiations of submodular mutual information (SMI) functions to obtain a balanced labeled set. Concretely, we formulate an SMI function using a query set $\Qcal^c$ that is a subset of the current labeled set $\Lcal$ with data points from class $c$. The SMI functions are instantiated using a similarity kernel $\Xcal$, where $X_{ij}$ is the pairwise similarity between data points $i$ and $j$, represented by gradients computed using $\Mcal$. Specifically, we define $\Xcal_{ij} = \langle \nabla_{\theta} \Hcal_{\Lcal_i}(\theta), \nabla_{\theta} \Hcal_{\Lcal_j}(\theta) \rangle$, where $\Hcal_{\Lcal_i}(\theta) = \Hcal_\Lcal(x_i, y_i, \theta)$ is the labeled loss on the $i$th data point. Note that we use hypothesized labels, \ie\ the label with maximum class probability, for computing gradients of $\Ucal$. Next, we optimize the SMI function for class $c$ using a greedy strategy \cite{mirzasoleiman2015lazier} with a constraint that the budget is equally divided across all classes: 

\vspace{-1.5ex}
\begin{align}\label{eq:perclass-smi}
\max_{\Acal_i^c \subseteq \Ucal, |\Acal_i^c| \leq (B/C)} I_f(\Acal_i^c; \Qcal^c)  
\end{align}

Hence, \model\ gradually builds a balanced labeled set $\Lcal$ by accumulating smaller balanced sets $A_i$ in every AL round. Finally, we can use any SSL technique to train the model $\Mcal$ with the balanced labeled set $\Lcal$ and the remaining unlabeled set $\Ucal \backslash \Lcal$. We summarize our method in \AlgRef{algo:basil}, illustrate the architecture in \figref{fig:basil_arch}, and discuss its scalability in \Appref{app:scalability}.

\section{Experiments} \label{sec:experiments}
In this section, we evaluate the effectiveness of \model\ on two modalities of medical data, {\em viz.}, histopathology (see \secref{sec:pathmnist_analysis}) using the Path-MNIST dataset \cite{kather2019predicting,medmnistv2} and Abdominal CT (see \secref{sec:organmnist_analysis}) using the Organ-MNIST dataset \cite{kermany2018identifying, medmnistv2}. For evaluation, we compare the test accuracy of the model obtained after training via a semi-supervised learning algorithm on a combination of labeled set (selected using active learning) and the unlabeled set. We also compare the imbalance ratio (IR) of all AL methods in \figref{fig:res_imbalance_ratio}. The IR is computed as follows: $IR(\Acal) = \frac{|\Acal_\Fcal| * |\Rcal|}{|\Acal_\Rcal| * |\Fcal|}$, where $\Rcal$ contains class indices of the rare classes and $\Fcal$ contains class indices of the remaining frequent classes. Note that IR($\Acal$)=1 when $\Acal$ is perfectly balanced.

 

Our results show that selecting a balanced labeled set using \model\ (see \figref{fig:res_imbalance_ratio}) outperforms existing AL baselines (supervised row in \tabref{tab:pathmnist_res}, 2). Importantly, we observe that, independent of the SSL algorithm used, using a balanced labeled set selected using \model\ helps leverage the remaining imbalanced unlabeled set better. \looseness-1

\noindent \textbf{Baseline AL methods and SSL techniques.} We compare the performance of \model\ against an uncertainty based AL method (\textsc{Entropy}), and a diversity based AL method (\textsc{Badge}). We discuss the details of all baselines in \secref{sec:related_work}. Lastly, we compare with random sampling (\textsc{Random}). We evaluate the subset selected by \model\ and the AL baselines on a wide array of SSL algorithms such as Psuedo-Label (PL), Mean Teacher (MT), $\Pi$-Model, MixMatch (MM), ICT, VAT, VAT + Entropy Minimization (EM) on two medical imaging datasets which are described in \secref{sec:pathmnist_analysis} and \secref{sec:organmnist_analysis}.

\noindent \textbf{Experimental setup:} 
We use the same training procedure and hyperparameters for all AL methods to ensure a fair comparison. For the first AL round, we randomly sample data points for labeling from the unlabeled set $\Ucal$. We do so in order to obtain meaningful model parameters for the AL acquisition functions in the next round of AL. For all experiments, we train a Wide ResNet (WRN)~\cite{he2016deep} model using an Adam optimizer with an initial learning rate of 3e-4. For each AL round, the weights are reinitialized using Xavier initialization and the model is trained for 100K iterations. After obtaining the labeled set using AL, we reinitialize the WRN model and train it using SSL for 500K iterations.  We run each experiment $3 \times$ on a V100 GPU and provide the error bars (std deviation). We discuss dataset splits and hyperparameters below and provide more details in \Appref{app:dataset_details}. \looseness-1

\vspace{-2ex}
\subsection{Analysis on Histopathology data} \label{sec:pathmnist_analysis}
\noindent\textbf{Path-MNIST Dataset:} Path-MNIST \cite{kather2019predicting} is a dataset based on a prior study for predicting survival from colorectal cancer histology slides. 
It includes a training dataset of 100,000 non-overlapping image patches from hematoxylin and eosin stained histological images, and a test dataset 
of 7,180 image patches from a different clinical center. These patches are categorized into 9 types of tissues, resulting in a multi-class classification task. For our experiments, we use a pre-processed version of Path-MNIST \cite{medmnistv2}, where the original images are resized to 3 $\times$ 28 $\times$ 28. We consider a subset of 26K data points to create the initial imbalanced unlabeled set $\Ucal$. The unlabeled set is imbalanced by randomly choosing 4 rare classes and randomly selecting 250 data points for each class. For the remaining 5 classes, we randomly select 5000 data points for each class. 


\vspace{-2ex}
\begin{table}[]
\small
\centering
\begin{small} 
\begin{tabular}{l|lll|lll}
\toprule
\textbf{SSL \textbackslash{} AL}                    & \textbf{\textsc{Random}}                          & \textbf{ \textsc{Badge}}                           & \textbf{\textsc{Entropy}}                         & \textbf{\textsc{Flqmi}}                                    & \textbf{\textsc{Gcmi}}                                     & \textbf{\textsc{Flvmi}}                                    \\
                    \midrule
\textbf{Supervised} & 69.37{\tiny $\pm$1.33} & 76.56{\tiny $\pm$0.4}  & 70.54{\tiny $\pm$1.76} & 78.51{\tiny $\pm$0.58}          & \textbf{78.76{\tiny $\pm$1.84}}          & 77.68{\tiny $\pm$0.58}          \\
\textbf{PL} \cite{lee2013pseudo}         & 69.38{\tiny $\pm$2.19} & 80.11{\tiny $\pm$1.68} & 66.16{\tiny $\pm$1.8}  & 78.66{\tiny $\pm$2.23}          & 80.33{\tiny $\pm$1.6}           & \textbf{81.24{\tiny $\pm$0.91}} \\
\textbf{ICT}  \cite{verma2019interpolation}      & 75.45{\tiny $\pm$0.28} & 80.25{\tiny $\pm$0.17} & 74.91{\tiny $\pm$0.78} & 82.29{\tiny $\pm$0.96}          & \textbf{82.58{\tiny $\pm$0.13}}          & 82.08{\tiny $\pm$0.66}          \\
\textbf{$\Pi$-Model} \cite{laine2016temporal}        & 64.59{\tiny $\pm$0.19} & 78.63{\tiny $\pm$1.03} & 67.29{\tiny $\pm$0.73} & 80.62{\tiny $\pm$1.72}          & 80.59{\tiny $\pm$1.19}          & \textbf{80.77{\tiny $\pm$0.71}} \\
\textbf{MT} \cite{tarvainen2017mean}         & 70.98{\tiny $\pm$1.76} & 79.62{\tiny $\pm$0.85} & 74.19{\tiny $\pm$1.56} & 82.14{\tiny $\pm$0.91}          & 82.43{\tiny $\pm$1.44}          & \textbf{83.36{\tiny $\pm$0.52}} \\
\textbf{MM} \cite{berthelot2019mixmatch}         & 67.13{\tiny $\pm$1.78} & 67.77{\tiny $\pm$1.43} & 67.01{\tiny $\pm$0.44} & 73.78{\tiny $\pm$1.22}          & \textbf{76.08{\tiny $\pm$0.09}} & 73.42{\tiny $\pm$1.54}          \\
\textbf{VAT} \cite{miyato2018virtual}       & 80.27{\tiny $\pm$1.95} & 83.05{\tiny $\pm$1.72} & 77.49{\tiny $\pm$1.8}  & \textbf{85.72{\tiny $\pm$1.83}} & 83.7{\tiny $\pm$0.37}           & 84.38{\tiny $\pm$0.81}          \\
\textbf{VAT+EM} \cite{miyato2018virtual}     & 81.48{\tiny $\pm$1.1}  & 84.11{\tiny $\pm$0.06} & 82.6{\tiny $\pm$0.75}  & \textbf{85.46{\tiny $\pm$1.54}} & 84.1{\tiny $\pm$1.94}           & 85.22{\tiny $\pm$1.51} \\
\bottomrule
\end{tabular}
\end{small}
\caption{Active SSL on Path-MNIST. Test accuracies in bold (one in each row) show the best performing AL acquisition functions for a particular SSL method. We observe that \textsc{Flqmi} which models query-relevance and representation obtains the highest gains for VAT and VAT+EM. Whereas, \textsc{Flvmi} acquires the best labeled set for PL, $\Pi$-Model and MT.}
\label{tab:pathmnist_res}
\end{table}
\vspace{-8ex}

\begin{figure*}[h]
\centering
\begin{subfigure}[]{0.48\textwidth}
\includegraphics[width = \textwidth]{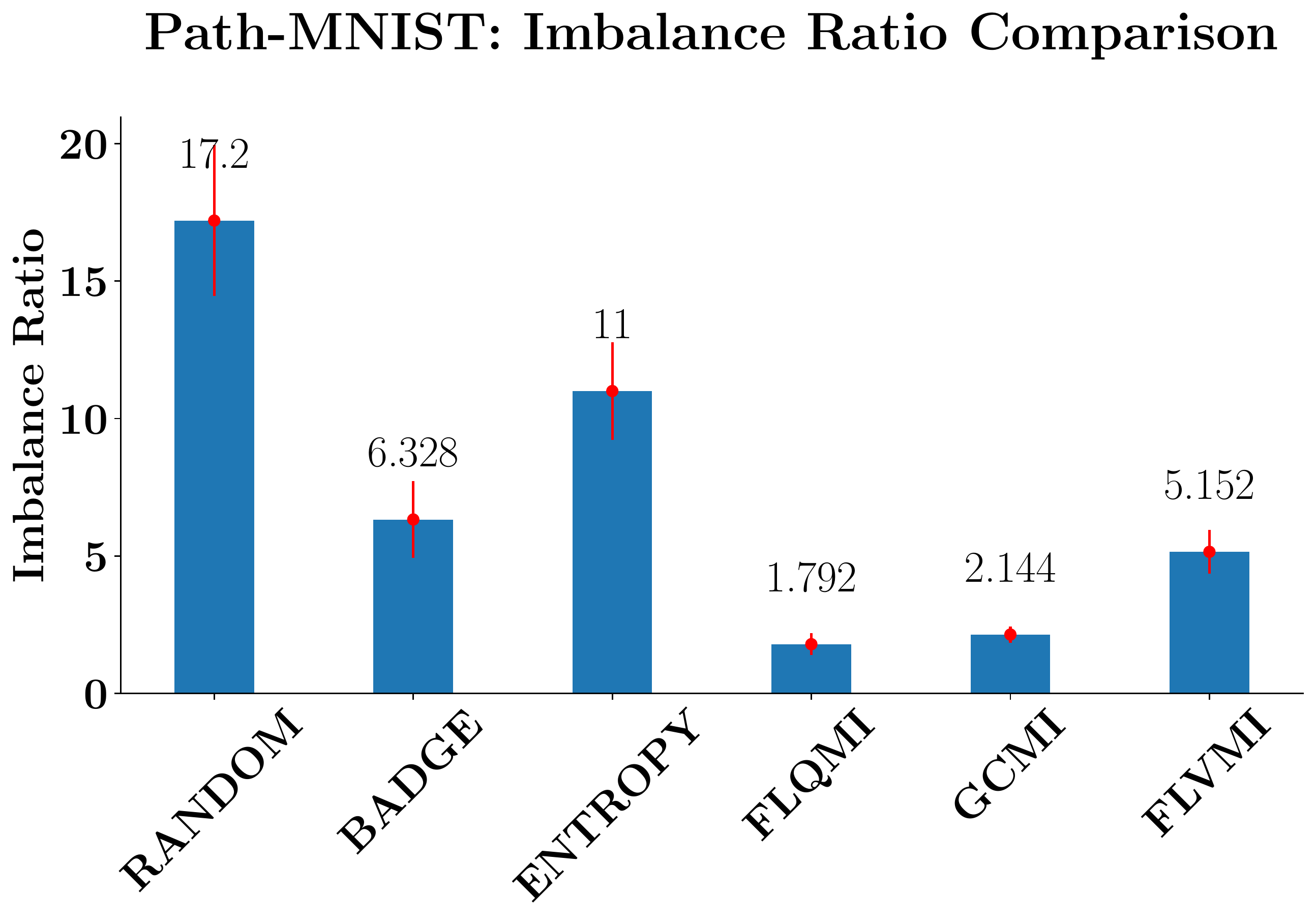}
\end{subfigure}
\begin{subfigure}[]{0.48\textwidth}
\includegraphics[width = \textwidth]{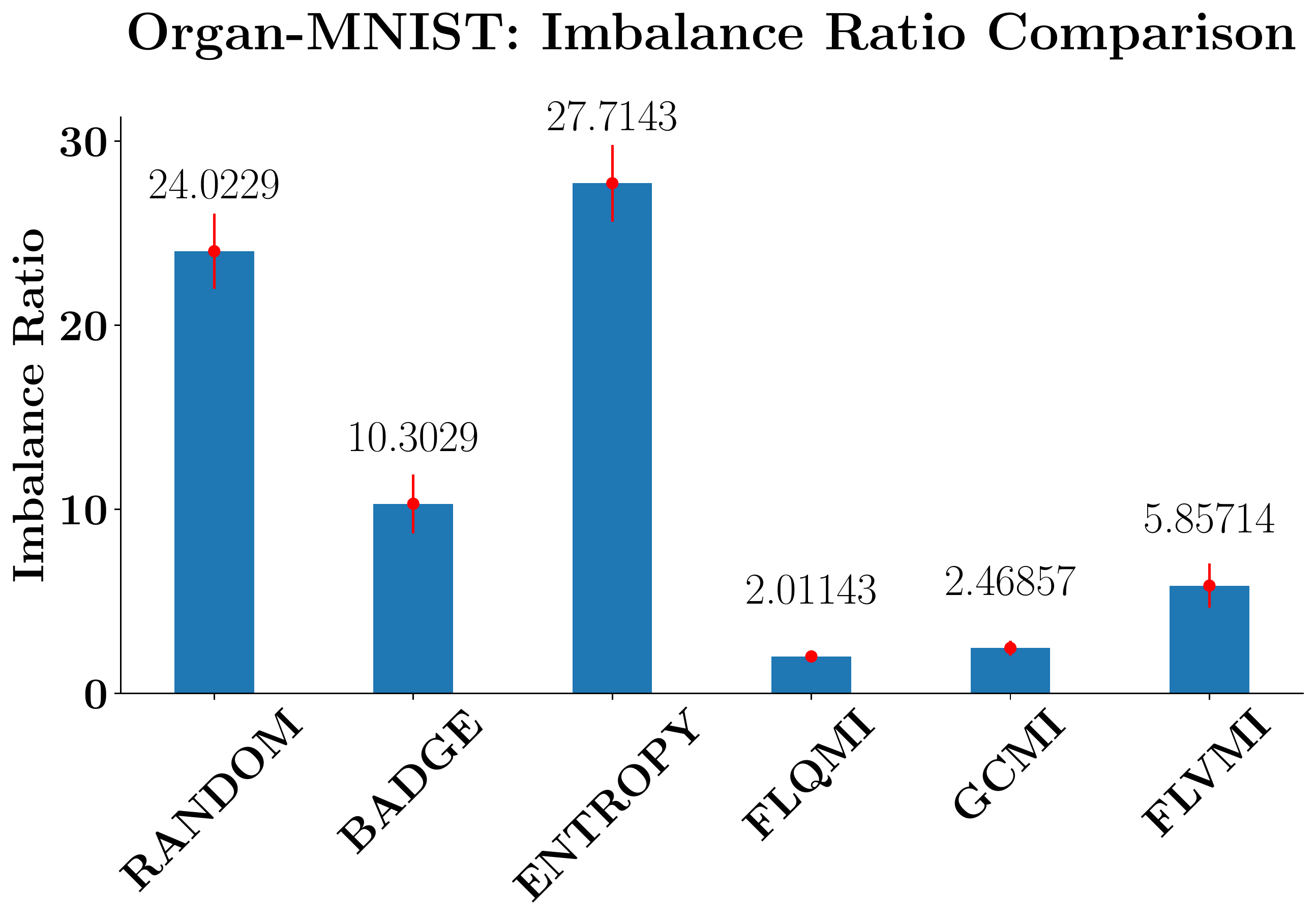}
\end{subfigure}
\caption{Comparison of Imbalance Ratio (IR) on Path-MNIST (Left) and Organ-MNIST (Right) datasets (Lower the better). We observe that the SMI functions select more data points from the rare classes, thereby having a $\approx 4\times$ to $5\times$ lower IR than the best performing baseline and $\approx 10\times$ to $20\times$ lower IR than \textsc{Random}.}
\label{fig:res_imbalance_ratio}
\end{figure*}
\vspace{-4ex}

\noindent\textbf{Results:} 
We present results for Active SSL on Path-MNIST in \tabref{tab:pathmnist_res}. We observe that the SMI based AL acquisition functions outperform existing AL methods in the supervised and across all semi-supervised learning methods. This is due to the fact that per-class selection using SMI functions in \model\ results in a more balanced labeled set (see \figref{fig:res_imbalance_ratio}). This reinforces the need for a framework like \model\ for training models in a supervised or semi-supervised manner in class imbalance scenarios. Interestingly, we observe that the choice of SMI function depends on the modality of medical data and the SSL method. For histopathology, we see that PL, $\Pi$-Model and MT methods perform the best when an acquisition function like \textsc{Flvmi} that balances between query-relevance and diversity is used. For ICT and MM, \textsc{Gcmi} shows the best results, which models query relevance only. Whereas, on VAT and VAT+EM, \textsc{Flqmi} shows the best results which models query-relevance and representation.

\subsection{Analysis on Abdominal CT data} \label{sec:organmnist_analysis}
\noindent \textbf{Organ-MNIST Dataset:} Organ-MNIST \cite{kermany2018identifying} is a dataset based on 3D computed tomography (CT) images from Liver Tumor Segmentation Benchmark (LiTS). For our experiments, we use a pre-processed version of Organ-MNIST \cite{medmnistv2}, where images are cropped using bounding-box annotations of 11 body organs. 
The images are resized into 1 $\times$ 28 $\times$ 28 to perform multi-class classification of 11 body organs. We consider a subset of 21.6K data points to create the initial imbalanced unlabeled set $\Ucal$. The unlabeled set is imbalanced by randomly choosing 4 classes and randomly selecting 150 data points for each class. For the remaining 5 classes, we randomly select 3000 data points for each class. \looseness-1

\noindent\textbf{Results:} 
We present results for Active SSL on Organ-MNIST in \tabref{tab:organmnist_res}. Similar to our results on Path-MNIST, we observe that the SMI based AL acquisition functions outperform existing AL methods across all supervised and SSL methods. We observe that the facility location based SMI functions dominate for the Abdominal CT modality. Particularly, \textsc{Flqmi} which models query-relevance and representation, outperforms other baselines and SMI functions for all SSL methods except VAT. The \textsc{Flvmi} functions performs slightly better than \textsc{Flqmi} when the SSL method is VAT or VAT+EM.

\begin{table}[]
\centering
\begin{tabular}{l|lll|lll}
\toprule
            \textbf{SSL \textbackslash{} AL}        & \textbf{\textsc{Random}}                          & \textbf{\textsc{Badge}}                        & \textbf{\textsc{Entropy}}                         & \textbf{\textsc{Flqmi}}                                    & \textbf{\textsc{Gcmi}}                            & \textbf{\textsc{Flvmi}}                                    \\
                    \midrule
\textbf{Supervised} & 60.63{\tiny $\pm$0.73} & 64.32{\tiny $\pm$1.29} & 61.03{\tiny $\pm$1.36} & \textbf{65.58{\tiny $\pm$0.28}} & 62.15{\tiny $\pm$1.63} & 64.41{\tiny $\pm$0.6}           \\
\textbf{PL} \cite{lee2013pseudo}        & 62.83{\tiny $\pm$1.79} & 64.43{\tiny $\pm$1.47} & 64.27{\tiny $\pm$0.16} & \textbf{67.68{\tiny $\pm$1.43}} & 64.48{\tiny $\pm$1.3}  & 65.56{\tiny $\pm$0.35}          \\
\textbf{ICT}  \cite{verma2019interpolation}      & 61.1{\tiny $\pm$0.57}  & 60.29{\tiny $\pm$1.52} & 64.01{\tiny $\pm$1.27} & \textbf{65.32{\tiny $\pm$0.86}} & 60.33{\tiny $\pm$1.37} & 62.22{\tiny $\pm$1.24}          \\
\textbf{$\Pi$-Model} \cite{laine2016temporal}         & 64.61{\tiny $\pm$1.32} & 65.61{\tiny $\pm$1.73} & 62.67{\tiny $\pm$1.83} & \textbf{65.94{\tiny $\pm$1.03}} & 61.85{\tiny $\pm$1.62} & 65.43{\tiny $\pm$0.94}          \\
\textbf{MT}  \cite{tarvainen2017mean}       & 64.64{\tiny $\pm$0.63} & 66.49{\tiny $\pm$1.18} & 65.53{\tiny $\pm$0.34} & \textbf{66.89{\tiny $\pm$0.01}} & 60.81{\tiny $\pm$0.63} & 64.81{\tiny $\pm$0.54}          \\
\textbf{MM}   \cite{berthelot2019mixmatch}      & 53.62{\tiny $\pm$1.28} & 50.34{\tiny $\pm$1.4}  & 51.35{\tiny $\pm$1.51} & \textbf{58.57{\tiny $\pm$0.51}} & 55.86{\tiny $\pm$0.73} & 56.08{\tiny $\pm$1.07}          \\
\textbf{VAT}  \cite{miyato2018virtual}      & 71.82{\tiny $\pm$1.98} & 70.48{\tiny $\pm$0.53} & 72.17{\tiny $\pm$0.97} & 75.25{\tiny $\pm$0.73}          & 72.51{\tiny $\pm$1.62} & \textbf{76.17{\tiny $\pm$0.1}}  \\
\textbf{VAT+EM} \cite{miyato2018virtual}     & 72.67{\tiny $\pm$0.28} & 73.52{\tiny $\pm$0.22} & 71.95{\tiny $\pm$0.43} & 75.57{\tiny $\pm$1.12}          & 73.17{\tiny $\pm$1.73} & \textbf{76.58{\tiny $\pm$0.69}} \\
\bottomrule
\end{tabular}
\caption{Active SSL on Organ-MNIST. Test accuracies in bold (one on each row) show the best performing AL method for each SSL. We see that \textsc{Flqmi} which models query-relevance and representation obtains the highest gains for most SSL methods except VAT, where \textsc{Flvmi} consistently performs better.}
\label{tab:organmnist_res}
\end{table}

\section{Conclusion}
We demonstrate the effectiveness of a unifying algorithm like \model\ for selecting a balanced labeled set in scenarios with class imbalanced data. Through rigorous experiments on diverse modalities of medical datasets, we show that \model\ selects a more balanced labeled set than other AL acquisition functions, resulting in relatively unbiased models leading to better performance for supervised and semi-supervised learning.

\bibliographystyle{splncs04}
\bibliography{main}

\newpage

\appendix

\setcounter{page}{1}

\section*{Supplementary Material} 

\section{Summary of Notations}\label{app:notation-summary}

 \begin{table*}[!h]
 \centering
 \begin{tabular}{|l|l|p{0.5\textwidth}|} 
 \toprule
 \hline 
 \multicolumn{1}{|l|}{Topic} & Notation & Explanation \\ \hline
 \toprule \hline
 &  $\Ucal$ & Unlabeled set of $|\Ucal|$ instances\\ 
 \multicolumn{1}{|p{0.20\textwidth}|}{\model\ (\secref{sec:our_method})} 
 & $\Acal$ & A subset of $\Ucal$\\ 
 & $S_{ij}$ & Similarity between any two data points $i$ and $j$\\
 & $f$ & A submodular function\\
 & $\Lcal$ & Labeled set of data points\\
 & $\Qcal$ & Query set\\
 & $\Qcal^c$ & Query set containing data points from class $c$ for per-class SMI selection, $\Qcal^c \subseteq \Lcal$\\
 & $\Mcal$ & Deep model\\
 & $B$ & Active learning selection budget\\
 & $\Hcal_\Lcal$ & Labeled Loss function used to train model $\Mcal$ and compute gradients\\
 & $\Hcal_\Ucal$ & Unlabeled Loss function used for semi-supervised learning\\
 & $\Xcal$ & Pairwise similarity matrix computed using gradients\\
 & $\Gcal_\Acal$ & Gradients of some subset $\Acal$\\
 \hline
 \bottomrule
 \end{tabular}
 \caption{Summary of notations used throughout this paper}
 \label{tab:main-notations}
 \end{table*}

\section{Details of Datasets and Experimental setting} \label{app:dataset_details}
\subsection{Datasets}
In this section, we will describe details of the datasets Path-MNIST (Histopathology data) and Organ-MNIST (Abdominal CT data).
\subsubsection{Path-MNIST}
A dataset based on a prior study for predicting survival from colorectal cancer histology slides, which provides 100,000 non-overlapping image patches from hematoxylin and eosin stained histological images, and a test dataset of 7,180 image patches from a different clinical center. 9 types of tissues are involved, resulting a multi-class classification task. We resize the source images of 3 x 224 x 224 into 3 x 28 x 28. \\
\begin{figure}[!h]
    \makebox[1 \textwidth][c]{
    \resizebox{\textwidth}{!}{
    \begin{tabular}{|l|l|l|l|l|l|l|l|l|}
    \toprule
    \textbf{0} & \textbf{1} & \textbf{2} & \textbf{3} & \textbf{4} & \textbf{5} & \textbf{6} & \textbf{7} & \textbf{8} \\
    \hline
    \textit{adipose} & \textit{background} & \textit{deris} & \textit{lymphocytes} & \textit{mucus} & \textit{smooth muscle} & \textit{normal colon mucus} & \textit{cancer-associated stroma} & \textit{colorectal adenocarcinoma epithelium}\\
    \bottomrule
    \end{tabular}
    }}
    \caption{Types of tissues and their labels in Path-MNIST}
    \label{tab:my_label}
\end{figure}
\\Out of 100K images we take 5K images each from classes {0,1,4,6,8} and 250 images each from classes {2,3,5,7} and form a unlabeled train dataset of 26K images and for validation set we had 10 samples each from all classes thus forming a validation set of size 90. For the test dataset we used default one which consists of 7180 samples. For Path-MNIST we had active learning selection budget($B$) as 900. 
\subsubsection{Organ-MNIST}
A dataset based on 3D computed tomography (CT) images from Liver Tumor Segmentation Benchmark (LiTS). Hounsfield-Unit (HU) of the 3D images are transformed into grey scale with a abdominal window; we then crop 2D images from the center slices of the 3D bounding boxes in axial views (planes). The images are resized into 1 x 28 x 28 to perform multi-class classification of 11 body organs. \\
\begin{figure}[!h]
    \makebox[1 \textwidth][c]{
    \resizebox{\textwidth}{!}{
    \begin{tabular}{|l|l|l|l|l|l|l|l|l|l|l|}
    \toprule
    \textbf{0} & \textbf{1} & \textbf{2} & \textbf{3} & \textbf{4} & \textbf{5} & \textbf{6} & \textbf{7} & \textbf{8} & \textbf{9} & \textbf{10} \\
    \hline
    \textit{bladder} & \textit{femur-left} & \textit{femur-right} & \textit{heart} & \textit{kidney-left} & \textit{kidney-right} & \textit{liver} & \textit{lung-left} & \textit{lung-right} & \textit{pancreas} & \textit{spleen}\\
    \bottomrule
    \end{tabular}
    }}
    \caption{Organs and their labels in OrganA-MNIST}
    \label{tab:my_label}
\end{figure}
There are total of 34581 training samples out of which we pick 3000 samples from classes {4,5,6,7,8,9,10} and 150 samples from classes {0,1,2,3} and form an unlabeled train dataset of 21.6K images. And for validation set we have selected 10 points each from a class to form a validation set of size 110. And the test set consists a total of around 17K images. For Organ-MNIST we had active learning selection budget($B$) as 990.
\subsection{Experimental Setting}
Given an unlabeled dataset $\Ucal$ and an active learning selection budget $B$, we have to select $B/N$ points in each round of active learning. In the first round, we select $B/N$ points randomly from the unlabeled dataset and label those points. Then in further rounds we train the Wide-ResNet model using an Adam optimizer with an initial learning rate of $3e-3$ (same across different active learning methods to make better comparison) for 100K iterations using the labeled set formed till then and extract the gradients of remaining unlabeled dataset from model and use the gradients to perform per-class selection from SMI functions to get a Balanced Dataset.\\
After forming a labeled set of size $B$, we perform Semi-Supervised learning using labeled set $\Lcal$ and unlabeled dataset $\Ucal$. We use many SSL algorithms to evaluate our selected balanced datasets such as Psuedo-Label, ICT, $\Pi$-Model, Mean-Teacher, MixMatch, VAT, VAT+EM. For each of these methods the loss component consists of two components, they are supervised loss (loss corresponding to labeled dataset) and SSL loss (loss corresponding to unlabeled dataset). Supervised loss is common to all, whereas SSL loss depends on the algorithm which we are using. We use the same parameters for one SSL method across different selections of active learning for better comparison.
\subsubsection{Parameters used for each SSL method}
\begin{itemize}
    \item \textbf{Supervised}: {"lr": $3e-3$}
    \item \textbf{PL}: {"threshold": 0.95, "lr": $3e-4$, "consistency-coefficient": 1} \\
    \item \textbf{ICT}: {"ema\_factor": 0.999, "lr": $4e-4$, "consistency-coefficient": 100, "alpha": 0.1} \\ 
    \item $\Pi$\textbf{-Model}: {"lr": $3e-4$, "consistency-coefficient": 20.0} \\ 
    \item \textbf{MT}: {"ema\_factor": 0.95, "lr": $4e-4$, "consistency-coefficient": 8} \\
    \item \textbf{MM}: {"lr": $3e-3$, "consistency-coefficient": 100, "alpha": 0.75, "T": 0.5, "K": 2} \\
    \item \textbf{VAT}: {"xi": $1e-6$, "lr": $3e-3$, "consistency-coefficient": 0.3, "eps": 6} \\
    \item \textbf{VAT+EM}: {"xi": $1e-6$, "lr": $3e-3$, "consistency-coefficient": 0.3, "eps": 6, "em": 0.06} \\ 
    
\end{itemize}

\textbf{Details on the computation of Imbalance Ratio: } The IR is computed as follows: $IR(\Acal) = \frac{|\Acal_\Fcal| * |\Rcal|}{|\Acal_\Rcal| * |\Fcal|}$, where $\Rcal$ contains class indices of the rare classes and $\Fcal$ contains class indices of the remaining frequent classes. For example, for a dataset with 5 total classes, assume it has 2 rare classes, $\Rcal = \{1,2\}$, then size of $|\Rcal|=2$. The remaining classes are frequent classes, $\Fcal = \{3,4,5\}$ and, $|\Fcal|=3$. Further, $\Acal_\Rcal$ denotes the set of data points that belong to the rare classes. Similarly, $\Acal_\Fcal$ denotes the set of data points that belong to the frequent classes. Note that IR($\Acal$)=1 when $\Acal$ is perfectly balanced.

\section{Scalability of \model} \label{app:scalability}

Below, we provide a detailed analysis of the complexity of creating and optimizing the different SMI functions. Denote $|\Xcal|$ as the size of set $\Xcal$. Also, let $|\Ucal| = n$ (the ground set size, which is the size of the unlabeled set in this case). 
\begin{itemize}
    \item \textbf{Facility Location: } We start with FLVMI. The complexity of creating the kernel matrix is $O(n^2)$. The complexity of optimizing it is $\tilde{O}(n^2)$ (using memoization)\footnote{$\tilde{O}$: Ignoring log-factors} if we use the stochastic greedy algorithm~\cite{mirzasoleiman2015lazier} and $O(n^2k)$ with the naive greedy algorithm. The overall complexity is $\tilde{O}(n^2)$.
    For FLQMI, the cost of creating the kernel matrix is $O(n|\Qcal|)$, and the cost of optimization is also $\tilde{O}(n|\Qcal|)$ (with naive greedy, it is $O(nB |\Qcal|)$). 
    \item \textbf{Graph-Cut: } For GCMI, we require a $O(n|\Qcal|)$ kernel matrix, and the complexity of the stochastic greedy algorithm is also $\tilde{O}(n|\Qcal|)$. 
\end{itemize}
We end with a few comments. First, most of the complexity analysis above is with the stochastic greedy algorithm~\cite{mirzasoleiman2015lazier}. If we use the naive or lazy greedy algorithm, the worst-case complexity is a factor $B$ larger. Secondly, we ignore log-factors in the complexity of stochastic greedy since the complexity is actually $O(n\log 1/\epsilon)$, which achieves an $1 - 1/e - \epsilon$ approximation.



\end{document}